\documentclass[journal]{IEEEtran}
\IEEEoverridecommandlockouts
\usepackage{cite}
\usepackage{amsmath,amssymb,amsfonts}
\usepackage{algorithmic}
\usepackage{graphicx}
\usepackage{float}
\usepackage[caption=false,font=footnotesize]{subfig}
\usepackage{textcomp}
\usepackage{xcolor}
\usepackage{svg}
\usepackage{placeins}
\usepackage[numbers,sort&compress]{natbib}
\usepackage{booktabs}
\usepackage{multirow}
\usepackage[normalem]{ulem}
\usepackage{dblfloatfix}  
\usepackage{algorithm}
\usepackage{algorithmic}

\usepackage{amsthm}


\definecolor{ForestGreen}{RGB}{34,139,34}
\definecolor{MediumBlue}{RGB}{0,0,205}      

\usepackage{amsthm}

\newtheoremstyle{upremark}%
  {\topsep}{\topsep}%
  {\normalfont}
  {}
  {\bfseries}
  {}
  { }
  {\thmname{#1}\thmnumber{ #2}}%
\theoremstyle{upremark}

\usepackage{pgfplots}
\usetikzlibrary{patterns,arrows,positioning,calc,fadings,shapes,decorations.markings}
\usepackage{tikz}
\usepackage{bm}
\def\BibTeX{{\rm B\kern-.05em{\sc i\kern-.025em b}\kern-.08em
    T\kern-.1667em\lower.7ex\hbox{E}\kern-.125emX}}

\newcommand{\SO}{\mathrm{SO}}
\newcommand{\SE}{\mathrm{SE}}

\newcommand{\so}{\mathfrak{so}}
\newcommand{\Exp}{\mathrm{Exp}}
\newcommand{\Log}{\mathrm{Log}}
\newcommand{\diag}{\mathrm{diag}}

\newcommand{\I}{\mathbf I}
\newcommand{\0}{\mathbf 0}
\newcommand{\R}{\mathbf R}
\newcommand{\p}{\mathbf p}
\renewcommand{\v}{\mathbf v}
\newcommand{\g}{\mathbf g}
\newcommand{\bg}{\mathbf b_{g}}
\newcommand{\ba}{\mathbf b_{a}}

\newcommand{\bP}{\mathbf P}
\newcommand{\bQ}{\mathbf Q}

\newcommand{\bG}{\mathbf G}

\newcommand{\bK}{\mathbf K}

\newcommand{\bPhi}{\boldsymbol\Phi}

\providecommand{\todo}[1]{\textcolor{red}{[TODO]}}

\begin{document}

\title{LBDU-VIO: Learned Bias Dynamics and Uncertainty for Visual-Inertial Odometry with Unreliable Vision}

\author{Qizhi Guo, Junning Lyu, Defu Lin, Shaoming He

\thanks{This work was supported by the National Natural Science Foundation of China under Grant 52302449.}%
\thanks{Authors' addresses: Qizhi Guo, Junning Lyu, Defu Lin, and Shaoming He are with the School of Aerospace Engineering and the Beijing Key Laboratory of UAV Autonomous Control, Beijing Institute of Technology, Beijing 100081, China, E-mail: ({qizhi.guo@bit.edu.cn}; {3120215027@bit.edu.cn}; {lindf@bit.edu.cn}; {shaoming.he@bit.edu.cn}). \textit{(Corresponding authors: Defu Lin; Shaoming He.)}}%

}

\maketitle 

\begin{abstract}

Visual-inertial odometry (VIO) for aerial robots relies on high rate inertial measurement unit (IMU) propagation between visual updates. 
However, conventional multi state constraint Kalman filters (MSCKFs) use random walk bias assumptions and fixed noise parameters, which can limit robustness when visual information is unreliable.
To address this problem, we propose LBDU-VIO, a learning-augmented MSCKF with learned continuous time bias dynamics and an IMU uncertainty model.
A neural ordinary differential equation (ODE) models continuous time bias dynamics to propagate the filter's bias states, replacing their random walk model.
The IMU uncertainty model predicts motion adaptive measurement noise covariances for covariance propagation.
Both models are trained with pose supervision without direct labels.
Experiments on real world EuRoC and TUM-VI benchmarks show lower errors than representative visual-inertial baselines,
including a 25.1\% reduction in mean relative position error compared with S-MSCKF on EuRoC sequences with 10s visual outage.
\end{abstract}


\section{Introduction}
\label{sec:introduction}

Visual-inertial odometry (VIO) is widely used in aerial robots, where cameras and inertial measurement units (IMUs) provide lightweight and complementary sensing for state estimation~\cite{xue2022gaussian,qin2018vinsmono}.
In a typical VIO pipeline, visual measurements correct accumulated drift at a relatively low rate, while high-rate IMU propagation maintains the state estimate between consecutive visual updates~\cite{lyu2022structure, zheng2024fast}.
This propagation becomes especially important when visual information is unreliable, such as under motion blur, low texture environments, or temporary camera outage.
In these cases, the estimator must rely more heavily on inertial measurements, and IMU modeling errors can quickly accumulate into large estimation drift~\cite{guo2025amo, hutchinson1973kalman,huang2022mems}.
Therefore, the reliability of a VIO system largely depends on how accurately the IMU bias is modeled and estimated.
\begin{figure}[!tbp]
\centering
\includegraphics[width=\columnwidth]{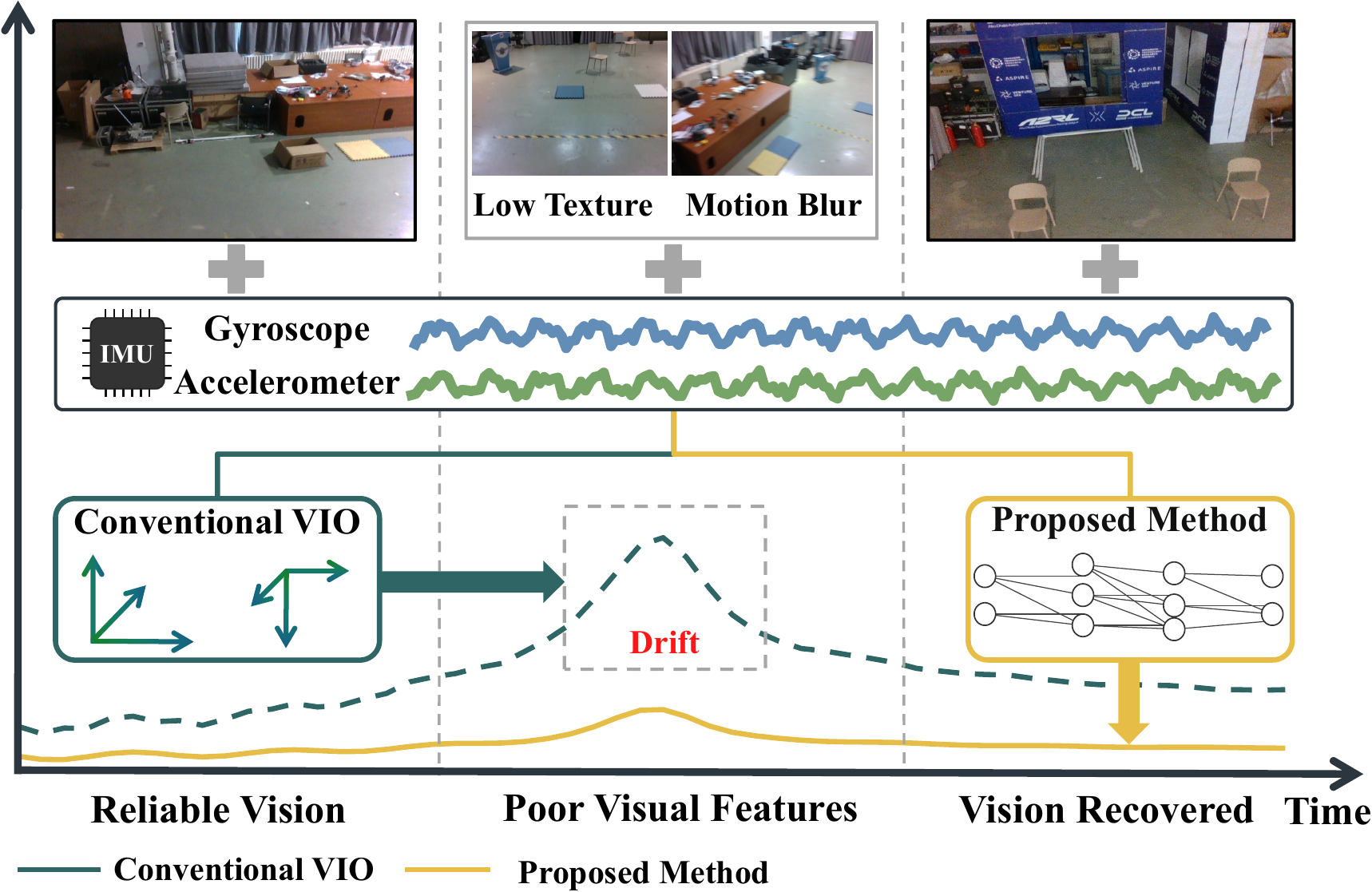}
\caption{Conceptual illustration of VIO under unreliable vision. The proposed LBDU-VIO aims to reduce drift during visual degradation and visual recovery after temporary tracking failure.
}
\label{fig:illustration}
\end{figure}

Conventional VIO estimators model gyroscope and accelerometer bias evolution as random walks with fixed noise parameters~\cite{ma2026small}.
This efficient formulation predicts a constant bias between measurement updates and uses prescribed noise statistics for covariance propagation~\cite{li2023cooperative}.
However, low cost IMUs often exhibit errors that are not captured by these assumptions~\cite{zhang2021low}.
The resulting model mismatch can affect both state prediction and uncertainty estimation.
Recent learning-based inertial methods predict discrete IMU corrections from fixed measurement windows~\cite{brossard2020denoising,buchanan2022deep}.
During UAV flight, biases vary with motion excitation within a window. A constant bias estimate can accumulate residual errors that increase inertial drift, while fixed noise covariances fail to capture motion dependent uncertainty.
This motivates combining learned continuous time bias dynamics with an IMU uncertainty model to improve inertial propagation when vision is unreliable.

To this end, we augment MSCKF with learned continuous time bias dynamics and an IMU uncertainty model to improve inertial propagation.
The resulting method termed LBDU-VIO targets the scenarios illustrated in Fig.~\ref{fig:illustration}.
A neural ordinary differential equation models the continuous time evolution of IMU biases, while an uncertainty model predicts motion adaptive measurement noise covariances. 
on. 
We evaluate LBDU-VIO on real world VIO benchmarks, including sequences with degraded images and periods without visual updates.
The main contributions are summarized as follows:
\begin{enumerate}

\item We develop LBDU-VIO, a learning-augmented MSCKF that couples learned continuous time IMU bias dynamics with visual updates.  Compared with assigning a single bias estimate for IMU measurements within a time window, our formulation models bias evolution to improve inertial propagation.

\item We introduce an IMU uncertainty model that predicts motion adaptive gyroscope and accelerometer measurement noise covariances from raw IMU data. These predictions replace fixed noise parameters in covariance propagation at every IMU sample.

\item We validate LBDU-VIO on real world EuRoC and TUM-VI benchmarks, and compare it with representative inertial and visual-inertial baselines. The results show improved state estimation under both nominal and unreliable visual conditions.
\end{enumerate}

The remainder of this paper is organized as follows. Section~\ref{sec:related_work} reviews related work. Section~\ref{sec:methodology} presents the continuous time IMU bias model and the IMU uncertainty model, along with their training and integration into MSCKF. Section~\ref{sec:experiments} reports experimental results, and Section~\ref{sec:conclusion} concludes the paper.

\section{Related Work}
\label{sec:related_work}
Learning-based IMU processing is increasingly applied to visual-inertial navigation. The field can be organized by what is learned and how it enters the estimator.
Early work learns a motion output directly from raw IMU sequences.
A complementary line targets the IMU error sources, while a further question is how to couple these learned quantities to the estimator.

\subsection{Learning Inertial Motion Models}
\label{subsec:rw_bias_learning}
Recent learning-based inertial methods take raw IMU sequences as the network input and regress motion directly. The motion refers to a kinematic quantity such as velocity, displacement, and orientation.

One group of methods predicts these motion states at each time step.
IONet~\cite{chen2018ionet} uses a recurrent network to map a fixed length inertial window to a displacement and heading change, bypassing the classical strapdown integration entirely.
RoNIN~\cite{herath2020ronin} regresses body frame velocity in a heading agnostic frame for pedestrian inertial navigation, while TLIO~\cite{liu2020tlio} extends this to a three-dimensional displacement increment together with its uncertainty.
AirIO~\cite{qiu2025airimu} also predicts body frame velocity with the current attitude as an additional network input. However, it regresses the motion output directly without modeling the underlying IMU bias, so the learned correction cannot be reused by the state estimator.
OriNet~\cite{esfahani2019orinet} targets the orientation channel alone, learning a drift-reduced orientation estimate from raw gyroscope sequences.
The second group of methods learns a relative motion increment over a finite window rather than a per step estimate.
RNIN-VIO~\cite{chen2021rnin} uses a recurrent network to predict a relative pose change and its uncertainty from a short IMU window. Learned IMU preintegration~\cite{cioffi2023learned} applies the same idea over a longer interval, replacing the classical preintegration of angular velocity and acceleration~\cite{forster2017manifold} with a learned method. Both methods fold an entire window into a single increment, so the sensor errors are absorbed into the output rather than modeled explicitly.

These methods span a range of motion outputs.
In every case, the network is supervised to reproduce a motion output rather than the underlying IMU error process. Such an end-to-end mapping generalizes poorly across motion regimes and is hard to interpret, since it never exposes the time-varying bias and measurement noise that cause inertial drift.
This motivates a complementary line of work that models the IMU bias and noise terms explicitly.

\subsection{Learning Inertial Bias and Noise Models}
\label{subsec:rw_uncertainty}

Inertial measurements enter a visual-inertial estimator through a fixed error model, in which the IMU biases follow random walks and the measurement noise is white Gaussian, both calibrated offline~\cite{li2023exploring, gao2025lm}.
This treatment underlies a broad range of aerial visual-inertial navigation systems, from interacting-multiple-model filters for micro aerial vehicles~\cite{gomaa2021observability} to sliding window estimators that fuse structural line priors for UAV inspection~\cite{lyu2022structure}.
The model is convenient for filter design but motion-invariant. It inflates the bias covariance at a fixed rate regardless of the actual excitation. It also keeps the measurement noise constant even though the noise of a low cost IMU varies with motion and vibration.

Currently, widely used learning-based methods address the IMU error as a single correction, without separating bias from measurement noise.
The network takes a sequence of raw inertial readings and outputs debiased measurements that are used directly for integration.
Brossard \emph{et al.}~\cite{brossard2020denoising} train a dilated convolutional network to denoise the raw gyroscope measurements for open loop attitude estimation.
However, it corrects only the gyroscope, leaving the accelerometer error and the translational drift it induces unaddressed.
Zhang \emph{et al.}~\cite{zhang2021imu} extend this idea to the full IMU, using a recurrent network to correct both the gyroscope and accelerometer readings before they enter a visual-inertial estimator.
Steinbrener~\emph{et~al.}~\cite{steinbrener2022improved} compare recurrent and attention-based architectures for the same denoising task, finding recurrent models more effective for high-rate inertial data.
Yuan \emph{et al.}~\cite{yuan2023imudb} further remove the need for ground-truth labels by introducing a self-supervised denoising objective evaluated on both the EuRoC and TUM-VI benchmarks.
These methods reduce integration drift, but the learned term is an implicit, window level correction that blurs the systematic bias together with the stochastic noise.
It leaves no separate bias state for the estimator to carry forward.

To recover such a state, several works model the IMU bias explicitly as a time varying term, distinct from the measurement noise.
Altawaitan \emph{et al.}~\cite{altawaitan2025learned} predict the bias at keyframe rate, while Yi \emph{et al.}~\cite{yi2026plug} learn a plug-and-play bias factor at the window level. In both cases the prediction is intermittent, so between updates the bias reverts to the random walk model.
Buchanan \emph{et al.}~\cite{buchanan2023deep} achieve continuous inference with a recurrent network whose hidden state persists across windows, but training requires ground truth bias labels from auxiliary sensor fusion.
Liu \emph{et al.}~\cite{liu2025debiasing} model the bias evolution in continuous time, producing a bias trajectory at the full IMU rate. However, their method focuses on bias correction, without learning to propagate uncertainty through the bias dynamics.
On the uncertainty side, AirIMU~\cite{qiu2024airimu} jointly predicts IMU corrections and per sample uncertainty for differentiable preintegration and covariance propagation, while TLIO~\cite{liu2020tlio} regresses displacement covariance at the window level.
AirIMU does not model continuous time dynamics for the filter's bias states, while our method integrates these dynamics with learned IMU uncertainty in MSCKF propagation.

\subsection{Estimator Integration of Learned Inertial Models}
\label{subsec:rw_integration}

Learned inertial models have usually been coupled to estimators through interfaces around, rather than inside the IMU propagation step.
At the preprocessing level, denoising~\cite{brossard2020denoising} and debiasing~\cite{liu2023duet} methods correct the raw IMU stream before it enters a conventional pipeline~\cite{geneva2020openvins, jiang2020dvio}. This refinement provides cleaner inertial inputs to the estimator, but does not modify the internal bias dynamics. Operationally, the learned module is integrated as an upstream measurement layer.
A tighter coupling is achieved at the factor level, where learned bias factor~\cite{yi2026plug} and learned preintegration~\cite{cioffi2023learned} explicitly incorporate the learned inertial term into the optimization objective. In this formulation, the learned prediction enters the factor graph as a constraint over the relevant keyframe states.
Nevertheless, such formulations introduce the learned inertial term only through keyframe level, rather than during each IMU integration interval.
Adaptive covariance methods~\cite{lin2024flm,yu2012adaptive} go one step further by adjusting the fusion weight itself, but they modify the measurement confidence rather than the inertial propagation model.
In all three cases the learned measurement remains outside the IMU rate integration, and the state evolution is still driven by a random walk bias model and fixed noise covariances \cite{lyu2024range}.

Our method instead incorporates both the learned bias dynamics and the learned IMU uncertainty model into the IMU rate state update.
The learned bias dynamics replace the filter's random walk bias model, while the predicted gyroscope and accelerometer noise covariances replace the fixed values in covariance propagation.
These learned inertial quantities therefore shape both the predicted mean and covariance before any visual update is applied.

\section{Problem Statement}
\label{sec:problem_statement}

Since direct use of visual measurements is vulnerable to visual degradation, a robust approach is to refine the underlying inertial propagation. We seek a model that enhances both the propagated state and its uncertainty. This section begins with the IMU kinematics and measurement models, and leads to a formulation of the estimation problem.

\subsection{Preliminaries}
\label{subsec:preliminaries}

Consider a UAV equipped with an IMU and a camera. The IMU supplies high rate state prediction, while visual features provide lower rate corrections.
The continuous time strapdown kinematics of the UAV are described by
\begin{equation}
\begin{aligned}
\dot{\R}(t)
&=\R(t)\boldsymbol\omega(t)^\times\\
\dot{\v}(t)
&=\R(t)\mathbf a(t)+\g\\
\dot{\p}(t)
&=\v(t)
\end{aligned}
\label{eq:nominal_dynamics}
\end{equation}
where the rotation matrix $\R(t)\in\SO(3)$ denotes the transformation from the body frame to the world frame. The velocity $\v(t)\in\mathbb R^3$ and position $\p(t)\in\mathbb R^3$ are expressed in the world frame. The gravity vector in the world frame is defined as $\g=\begin{bmatrix}0&0&-g\end{bmatrix}^T$. The notation $(\cdot)^\times:\mathbb R^3\rightarrow\so(3)$ represents the skew symmetric operator.
These dynamics are driven by the angular velocity $\boldsymbol\omega(t)$ and specific force $\mathbf a(t)$, both expressed in the body frame. The corresponding IMU measurements are modeled as
\begin{equation}
\begin{aligned}
\tilde{\boldsymbol\omega}(t)
&=\boldsymbol\omega(t)+\bg(t)+\mathbf n_g(t)\\
\tilde{\mathbf a}(t)
&=\mathbf a(t)+\ba(t)+\mathbf n_a(t)
\end{aligned}
\label{eq:meas_model}
\end{equation}
Here, $\tilde{\boldsymbol\omega}(t)$ and $\tilde{\mathbf a}(t)$ denote the corresponding IMU measurements, $\bg(t)$ and $\ba(t)$ represent the gyroscope and accelerometer biases, respectively. Furthermore, the associated measurement noises, denoted by $\mathbf n_g(t)$ and $\mathbf n_a(t)$, are zero-mean with covariance matrices $\bQ_{ng}$ and $\bQ_{na}$.
We collect the orientation, velocity, and position in the navigation state
\begin{equation}
\mathbf X(t)=
\begin{bmatrix}
\R(t) & \v(t) & \p(t) \\
\mathbf 0 & 1 & 0 \\
\mathbf 0 & 0 & 1
\end{bmatrix}
\in \SE_2(3)
\label{eq:state_SE23}
\end{equation}
Augmenting this navigation state with the IMU biases gives the current IMU state used for inertial propagation
\begin{equation}
\mathbf x(t) = \bigl(\mathbf X(t),\mathbf b(t)\bigr)
\in \SE_2(3)\times\mathbb R^6
\label{eq:state_def}
\end{equation}
where $\mathbf b(t)= \bigl[\bg(t)^T,\ba(t)^T\bigr]^T$ stacks the gyroscope and accelerometer biases.
Propagating $\mathbf x(t)$ requires both the navigation dynamics in \eqref{eq:nominal_dynamics} and a model for the evolution of $\mathbf b(t)$.

\begin{figure*}[!t]
\centering
\includegraphics[width=0.92\textwidth]{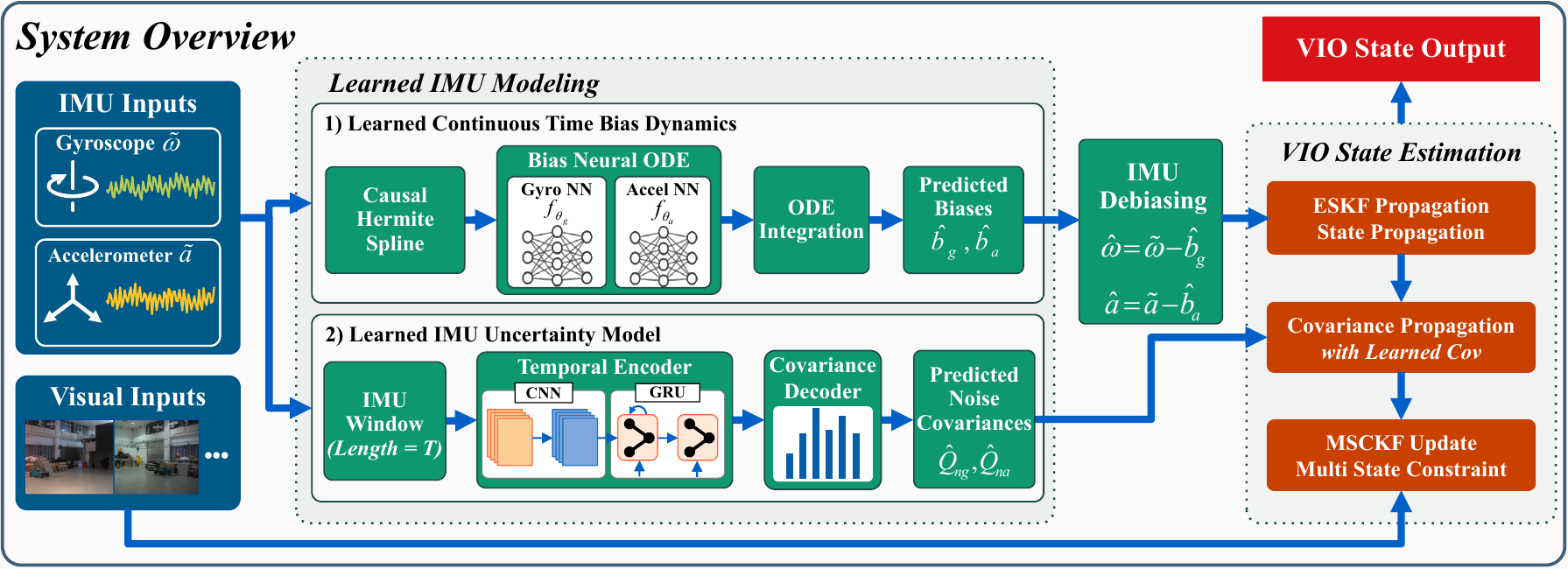}
\caption{LBDU-VIO system pipeline. A learning based multi state constraint Kalman filter that combines learned IMU bias dynamics and uncertainty with visual updates. Instead of modeling IMU biases as random walks and using fixed noise parameters, LBDU-VIO corrects raw IMU measurements with bias estimates from a neural ODE and uses measurement noise covariances predicted by an uncertainty model in covariance propagation.}
\label{fig:system_overview}
\end{figure*}
\subsection{Problem Formulation}

Our overall goal is to estimate the navigation state from IMU measurements and visual observations, including periods of unavailable vision. To support this goal, we combine learned IMU bias dynamics with an IMU uncertainty model for state and covariance propagation.
Methods that assign a single bias estimate to each time window may fail to capture the complex behavior of a low cost IMU, particularly bias variations within the window.
We therefore model the gyroscope and accelerometer biases as deterministic state variables governed by learned continuous time dynamics
\begin{equation}
\dot{\mathbf b}(t)
=\mathbf f\!\left(\mathbf b(t),\tilde{\mathbf u}(t),\dot{\tilde{\mathbf u}}(t)\right)
\label{eq:controlled_bias_dynamics}
\end{equation}
where $\mathbf f$ is a learned nonlinear vector field describing IMU bias evolution.
For notational convenience, let $\tilde{\mathbf u}(t)\triangleq\bigl[\tilde{\boldsymbol\omega}(t)^T,\tilde{\mathbf a}(t)^T\bigr]^T$ denote the measured IMU input.
The input derivative $\dot{\tilde{\mathbf u}}(t)$ is obtained from the interpolated IMU measurements.
IMU uncertainty model is constructed separately through the covariance matrices $\bQ_{ng}$ and $\bQ_{na}$ associated with $\mathbf n_g$ and $\mathbf n_a$ in \eqref{eq:meas_model}. 
Accordingly, we seek to estimate the IMU state through inertial propagation and visual updates using these learned models.

\textbf{Problem 1:}
Given an initial state estimate, use IMU measurements and available visual observations to estimate the current IMU state as
\begin{equation}
\hat{\mathbf x}_k
=\mathcal F\!\left(\tilde{\mathbf u}_{0:k},\mathcal Z_{\le t_k}\right)
\label{eq:prob_estimator}
\end{equation}
where $\hat{\mathbf x}_k$ estimates the IMU state defined in \eqref{eq:state_def}, and $\mathcal Z_{\le t_k}$ contains the visual observations available up to $t_k$.
The estimator $\mathcal F$ combines learned IMU bias dynamics and an IMU uncertainty model within MSCKF for state and covariance propagation under unreliable visual observations.


\section{Methodology}
\label{sec:methodology}

LBDU-VIO implements the estimator in \eqref{eq:prob_estimator} by integrating the learned models into MSCKF, as shown in Fig.~\ref{fig:system_overview}.
The filter propagates its state and covariance from raw gyroscope and accelerometer measurements using learned bias dynamics and adaptive measurement noise covariances.
It then uses visual measurements for multi state constraint updates.

\subsection{Learned Continuous Time Bias Dynamics}
\label{subsec:learned_bias_drift}

\begin{figure}[!hb]
\centering
\includegraphics[width=\columnwidth]{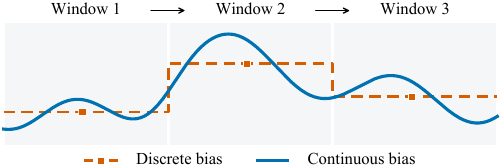}
\caption{Schematic comparison of bias representations. Learned continuous time dynamics describe bias evolution within each window, whereas a single estimate per window gives a piecewise constant bias.}
\label{fig:continuous_bias_schematic}
\end{figure}

We parameterize the vector field in \eqref{eq:controlled_bias_dynamics} with a neural network $\mathbf f_{\theta}$, where $\theta$ denotes its trainable parameters.
This formulation learns continuous time bias dynamics to capture variation within each window, rather than assigning a single constant estimate, as shown in Fig.~\ref{fig:continuous_bias_schematic}.
A neural ordinary differential equation models continuous state dynamics by parameterizing the right-hand side of an ODE with a neural network. For a generic state $\boldsymbol{\chi}(t)$, the canonical autonomous form is
\begin{equation}
\dot{\boldsymbol{\chi}}(t)=\mathbf f_{\vartheta}\!\left(\boldsymbol{\chi}(t)\right), \quad
\boldsymbol{\chi}(t_T)=\boldsymbol{\chi}(t_0)+
\int_{t_0}^{t_T}\mathbf f_{\vartheta}\!\left(\boldsymbol{\chi}(t)\right)dt
\label{eq:canonical_node}
\end{equation}
Here, $\mathbf f_{\vartheta}$ is a neural network with trainable parameters $\vartheta$ that maps the current state $\boldsymbol{\chi}(t)$ to its time derivative $\dot{\boldsymbol{\chi}}(t)$, and $t$ is the integration variable over $[t_0,t_T]$. 
In our setting, the bias serves as the neural ODE state, while its vector field also depends on the external IMU input.
This input is sampled at discrete times, it must be interpolated for the ODE solver to evaluate the dynamics between samples.
Therefore, we use a causal cubic Hermite spline $\mathbf S_{\tilde u}(t)$ to interpolate $\{\tilde{\mathbf u}_k\}_{k=0}^{N}$, with $\mathbf S_{\tilde u}(t_k)=\tilde{\mathbf u}_k$.
We model the stacked bias dynamics using the spline and its derivative
\begin{equation}
\dot{\mathbf b}(t)
=\mathbf f_{\theta}\!\left(
\mathbf b(t),
\mathbf S_{\tilde u}(t),
\dot{\mathbf S}_{\tilde u}(t)
\right)
\label{eq:bias_NODE_mean}
\end{equation}
where $\mathbf f_{\theta}$ stacks the gyroscope and accelerometer vector fields $\mathbf f_g$ and $\mathbf f_a$, which are separate residual multilayer perceptrons.
We learn the parameters $\theta$ from integrated motion errors, as detailed in Section~\ref{subsec:training}.
The spline derivative captures how the IMU measurements change over time, providing local temporal context for predicting bias evolution.
We set $\dot{\mathbf S}_{\tilde u}(t_0)=0$ and compute subsequent knot slopes using backward differences.
Integrating the bias dynamics in \eqref{eq:bias_NODE_mean} over $[t_0,t_T]$ produces the continuous bias trajectory
\begin{equation}
\begin{aligned}
\mathbf b(t_T)
&=\mathbf b_0+
\int_{t_0}^{t_T}
\mathbf f_{\theta}\!\left(
\mathbf b(t),t
\right)dt\\
&=\operatorname{ODESolve}\!\left(
\mathbf b_0,\mathbf f_{\theta},t_0,t_T;\theta
\right)
\end{aligned}
\label{eq:bias_integral_solution}
\end{equation}
Here, $\mathbf f_{\theta}(\mathbf b,t)$ is shorthand for the vector field in \eqref{eq:bias_NODE_mean}, with the spline and its derivative evaluated at time $t$.
The ODE solver obtains this continuous trajectory by numerically integrating the learned dynamics. In our implementation, we use explicit Euler integration at the IMU rate
\begin{equation}
\mathbf b_{k+1}
=\mathbf b_{k}
+\mathbf f_{\theta}\!\left(
\mathbf b_{k},t_k
\right)\Delta t_k
\label{eq:bias_euler}
\end{equation}
which is a numerical approximation of \eqref{eq:bias_integral_solution}. The sampling interval is \(\Delta t_k=t_{k+1}-t_k\).
At each IMU sample, we subtract the estimated biases from the raw measurements to obtain $\hat{\boldsymbol\omega}_k=\tilde{\boldsymbol\omega}_k-\hat{\mathbf b}_{g,k}$ and $\hat{\mathbf a}_k=\tilde{\mathbf a}_k-\hat{\mathbf b}_{a,k}$.
The bias and state dynamics are integrated jointly, yielding one updated bias estimate per IMU sample during online inference.

\subsection{Learned IMU Uncertainty Model and Differentiable Covariance Propagation}
\label{subsec:uncertainty_model}

To estimate gyroscope and accelerometer noise covariances, we encode a local IMU measurement window into a shared feature $\mathbf h_k=\mathcal E(\tilde{\mathbf U}_k)$ for both sensors.
Here, $\tilde{\mathbf U}_k$ denotes the local window of raw IMU measurements $\tilde{\mathbf u}$ associated with sample $k$, and $\mathcal E$ is the trainable temporal encoder.
The encoder consists of a convolutional front end followed by two gated recurrent unit layers.
The shared feature is then decoded into gyroscope and accelerometer noise covariance estimates at each IMU sample
\begin{equation}
\begin{aligned}
\bQ_{ng,k}&=\diag\!\left(\exp\!\left(\mathcal D_g(\mathbf h_k)\right)\right)\\
\bQ_{na,k}&=\diag\!\left(\exp\!\left(\mathcal D_a(\mathbf h_k)\right)\right)
\end{aligned}
\label{eq:Q_meas_learned}
\end{equation}
where $\bQ_{ng,k}$ and $\bQ_{na,k}$ are the continuous time noise covariance matrices for the gyroscope and accelerometer, respectively. 
The decoders $\mathcal D_g$ and $\mathcal D_a$ are separate multilayer perceptrons mapping $\mathbf h_k$ to three axis noise variances.
The $\exp$ operation maps the outputs of the trainable decoders $\mathcal D_g$ and $\mathcal D_a$ to positive values, and $\diag$ assembles these values into diagonal covariance matrices.
We supervise the uncertainty model with motion errors through differentiable preintegration and covariance propagation. Assuming IMU inputs keep constant within each sampling interval, the preintegration updates are
\begin{equation}
\begin{aligned}
\Delta\mathbf R_{k+1}
&=\Delta\mathbf R_k\operatorname{Exp}
\!\left(\hat{\boldsymbol\omega}_k\Delta t_k\right)\\
\Delta\mathbf v_{k+1}
&=\Delta\mathbf v_k
+\Delta\mathbf R_k\bar{\mathbf a}_k\Delta t_k\\
\Delta\mathbf p_{k+1}
&=\Delta\mathbf p_k+\Delta\mathbf v_k\Delta t_k
+\tfrac12\Delta\mathbf R_k\bar{\mathbf a}_k\Delta t_k^2
\end{aligned}
\label{eq:integrated_motion}
\end{equation}
Here, \(\Delta\mathbf R_k\), \(\Delta\mathbf v_k\), and \(\Delta\mathbf p_k\) denote the rotation, velocity, and position increments accumulated from the fixed window start \(t_i\) to the current time \(t_k\). They are expressed in the body frame at \(t_i\). 
The estimated linear acceleration is \(\bar{\mathbf a}_k=\hat{\mathbf a}_k+\hat{\mathbf R}_k^T\mathbf g\). The current orientation \(\hat{\mathbf R}_k=\hat{\mathbf R}_i\Delta\mathbf R_k\) is obtained by composing the initial orientation \(\hat{\mathbf R}_i\) with the accumulated relative rotation \(\Delta\mathbf R_k\).
The increments are initialized as \(\Delta\mathbf R_i=\I_3\), \(\Delta\mathbf v_i=\0_3\), and \(\Delta\mathbf p_i=\0_3\).
To quantify the uncertainty in these motion increments, we propagate the associated error covariance at each IMU step as
\begin{equation}
\boldsymbol\Sigma_{k+1}
=\mathbf A_k\boldsymbol\Sigma_k\mathbf A_k^T
+\mathbf B_k\mathbf Q_k\mathbf B_k^T
\label{eq:differentiable_covariance}
\end{equation}
Here, $\boldsymbol\Sigma_k$ is the IMU preintegration error covariance, initialized as $\boldsymbol\Sigma_i=\mathbf 0_{9\times 9}$. The accumulated preintegration error is modeled as $\boldsymbol\xi_k=[\delta\boldsymbol\phi_k^T,\delta\mathbf v_k^T,\delta\mathbf p_k^T]^T \sim \mathcal N(\mathbf 0_{9\times 1},\boldsymbol\Sigma_k)$, where $\delta\boldsymbol\phi_k$, $\delta\mathbf v_k$, and $\delta\mathbf p_k$ denote the zero mean rotation, velocity, and position errors, respectively.
The matrix $\mathbf A_k$ is the linearized error transition matrix, and $\mathbf B_k=[\mathbf B_{g,k}\ \mathbf B_{a,k}]$ is the noise injection matrix. The discrete measurement noise covariance is $\mathbf Q_k=\diag(\bQ_{ng,k},\bQ_{na,k})/\Delta t_k$.
Since the gravity term does not contribute to the attitude error Jacobian, the transition matrix in \eqref{eq:differentiable_covariance} is
\begin{equation}
\mathbf A_k=
\begin{bmatrix}
\Delta\R_{k,k+1}^T & \0 & \0\\
-\Delta\R_k\hat{\mathbf a}_k^{\times}\Delta t_k & \I & \0\\
-\tfrac{1}{2}\Delta\R_k\hat{\mathbf a}_k^{\times}\Delta t_k^2
& \I\Delta t_k & \I
\end{bmatrix}
\label{eq:preintegration_transition}
\end{equation}
Here, $\Delta\R_{k,k+1}=\Exp(\hat{\boldsymbol\omega}_k\Delta t_k)$ is the rotation increment from $t_k$ to $t_{k+1}$.
The gyroscope and accelerometer components of $\mathbf B_k$ are
\begin{equation}
\mathbf B_{g,k}
=
\begin{bmatrix}
\mathbf J_r^k\Delta t_k\\
\0\\
\0
\end{bmatrix}, \quad
\mathbf B_{a,k}
=
\begin{bmatrix}
\0\\
\Delta\R_k\Delta t_k\\
\tfrac{1}{2}\Delta\R_k\Delta t_k^2
\end{bmatrix}
\label{eq:preintegration_noise_injection}
\end{equation}
where $\mathbf J_r^k=\mathbf J_r(\hat{\boldsymbol\omega}_k\Delta t_k)$ denotes the $\mathrm{SO}(3)$ right Jacobian evaluated at the nominal rotation increment at time step $k$.

\subsection{Hierarchical Training Strategy}
\label{subsec:training}
We train our model in three stages using integrated motion residuals for supervision.
The rotation residual is $\mathbf e_{R,k}=\Log\,(\R_k^T\hat{\R}_k)$, where $\Log$ maps a rotation in the Lie group to the vector representation of Lie algebra in $\mathbb R^3$.
The velocity and position residuals are $\mathbf e_{v,k}=\R_i^T(\hat{\v}_k-\v_k)$ and $\mathbf e_{p,k}=\R_i^T(\hat{\p}_k-\p_k)$, respectively.
Here, $\R_i^T$ expresses both residuals from the world frame to the body frame at the window start~$t_i$.
\subsubsection{Stage~1: Gyroscope Bias Dynamics}
We train the gyroscope bias dynamics model $\mathbf f_g$ using orientation supervision, with the pose supervision mean squared error (MSE) and regularized objective defined as
\begin{equation}
\mathcal L_R
=\frac{1}{N}\sum_{k=1}^{N}
\left\|\mathbf e_{R,k}\right\|_2^2 + \lambda_g\mathcal L_{\mathrm{curv}}^{g}
\label{eq:stage_loss_rotation}
\end{equation}
Here, $k=1,\ldots,N$ indexes the $N$ supervised states along the integrated trajectory.
The coefficient $\lambda_g$ controls the strength of the gyroscope bias curvature regularization term.
This regularization is defined for both bias types as
\begin{equation}
\mathcal L_{\mathrm{curv}}^{i}
=\frac{1}{N-1}\sum_{k=1}^{N-1}
\left\|\dot{\mathbf b}_{i, k+1}-\dot{\mathbf b}_{i, k}\right\|_2^2, \quad i\in\{g,a\}
\label{eq:bias_curvature_regularization}
\end{equation}
It penalizes differences between consecutive bias derivatives to discourage unnecessarily rapid variations in the learned dynamics.

\subsubsection{Stage~2: Accelerometer Bias Dynamics}
We freeze $\mathbf f_g$ and train the accelerometer bias dynamics model $\mathbf f_a$ using velocity and position supervision. We also use an MSE loss and regularized objective
\begin{equation}
\mathcal L_{v,p}
=\frac{1}{N}\sum_{k=1}^{N}
\left(
\left\|\mathbf e_{v,k}\right\|_2^2
+\left\|\mathbf e_{p,k}\right\|_2^2
\right) + \lambda_a\mathcal L_{\mathrm{curv}}^{a}
\label{eq:stage_loss_velocity_position}
\end{equation}
where $\mathcal L_{\mathrm{curv}}^{a}$ is the accelerometer bias curvature regularization term defined in \eqref{eq:bias_curvature_regularization} with $i=a$, and $\lambda_a$ is its weight.
The derivatives defined in \eqref{eq:bias_NODE_mean} are jointly integrated with the inertial dynamics, allowing the loss functions to supervise the resulting motion without ground truth bias labels.
\subsubsection{Stage~3: IMU Uncertainty Model}
We freeze both bias models and train the uncertainty model on supervised IMU segments.
We use the mean negative log likelihood as the uncertainty loss
\begin{equation}
\mathcal L_{\mathrm{cov}}
=
\frac{1}{2N}
\sum_{k=1}^{N}
\sum_{c}
\left(
\mathbf e_{c,k}^T
(\boldsymbol\Sigma_k^c)^{-1}
\mathbf e_{c,k}
+
\ln(\det\boldsymbol\Sigma_k^c)
\right)
\label{eq:loss_uncertainty}
\end{equation}
where $c\in\{R,v,p\}$ indexes the rotation, velocity, and position components.
At each supervised time $t_k$, we sum the component losses over $c$ and then average the resulting losses over $k$.
The diagonal matrix $\boldsymbol\Sigma_k^c$ contains the corresponding variances from the preintegration error covariance $\boldsymbol\Sigma_k$ propagated by \eqref{eq:differentiable_covariance}.
The quadratic term weights the residuals by their inverse covariance, while the log determinant term penalizes large predicted variances.
We backpropagate this loss through the covariance recursion to update the temporal encoder and covariance decoders.

\subsection{LBDU-VIO Formulation}
\label{subsec:eskf}

We integrate the learned bias dynamics and IMU uncertainty model into an error state MSCKF. For the IMU state \(\mathbf x(t)\) defined in \eqref{eq:state_def}, the filter uses the local error
$
\delta\mathbf x=
[\delta\boldsymbol\theta^T,\delta\v^T,\delta\p^T,
\delta\bg^T,\delta\ba^T]^T\in\mathbb R^{15}
$, 
with \(\R=\hat{\R}\Exp(\delta\boldsymbol\theta^\times)\) and additive errors for the Euclidean components. 
Linearizing the error dynamics about the current state estimate yields the error model
\begin{equation}
\dot{\delta\mathbf x}(t)=
\mathbf F(t)\delta\mathbf x(t)+\bG(t)\mathbf n(t)
\label{eq:continuous_error_state_dynamics}
\end{equation}
Here, $\mathbf n=[\mathbf n_g^T,\mathbf n_a^T,\mathbf w_{bg}^T,\mathbf w_{ba}^T]^T$ stacks the measurement and bias process noises. The state and noise Jacobians are $\mathbf F(t)=\partial\dot{\delta\mathbf x}/\partial\delta\mathbf x$ and $\bG(t)=\partial\dot{\delta\mathbf x}/\partial\mathbf n$. They are computed from the nonlinear error dynamics at zero error and noise.
The learned dynamics defined in \eqref{eq:bias_NODE_mean} replace the bias random walks.
The corresponding gyroscope and accelerometer blocks of $\mathbf F$ are
\begin{equation}
\mathbf F_{bg}(t)=
\left.\frac{\partial\mathbf f_g}{\partial\mathbf b_g}\right|_{\hat{\mathbf b}_g}, \quad
\mathbf F_{ba}(t)=
\left.\frac{\partial\mathbf f_a}{\partial\mathbf b_a}\right|_{\hat{\mathbf b}_a}
\label{eq:filter_bias_dynamics}
\end{equation}
The filter propagates its bias estimate according to \eqref{eq:bias_euler}. 
The corrected IMU measurements drive the navigation dynamics in \eqref{eq:nominal_dynamics} to propagate the attitude, velocity, and position estimates.
The error state covariance is propagated between consecutive IMU samples as
\begin{equation}
\begin{aligned}
  \bP_{k+1}^{-}&=\bPhi_k\bP_k^{+}\bPhi_k^T+\bQ_{d,k}\\
\bPhi_k&=\exp(\mathbf F_k\Delta t_k)\\
\bQ_{d,k}&=\bG_k\bQ_{c,k}\bG_k^T\Delta t_k
\end{aligned}
\label{eq:P_discrete}
\end{equation}
where $\bP$ is the error state covariance, $\bPhi_k$ is the state transition matrix, and $\bQ_{d,k}$ is the discrete process noise covariance. 
The continuous time process noise covariance matrix is $\bQ_{c,k}=\operatorname{blkdiag}(\bQ_{ng,k},\bQ_{na,k},\bQ_{bg},\bQ_{ba})$. The uncertainty model provides $\bQ_{ng,k}$ and $\bQ_{na,k}$ according to \eqref{eq:Q_meas_learned}, while $\bQ_{bg}$ and $\bQ_{ba}$ are fixed bias process noise covariances. 
We apply the MSCKF visual update~\cite{geneva2020openvins} to the IMU state as
\begin{equation}
\begin{aligned}
\widehat{\delta\mathbf x}_k&=\bK_k\mathbf r_k\\
\hat{\mathbf x}_k^{+}&=\hat{\mathbf x}_k^{-}\boxplus\widehat{\delta\mathbf x}_k\\
\bP_k^{+}&=\bP_k^{-}-\bK_k\mathbf S_k\bK_k^T
\end{aligned}
\label{eq:visual_update}
\end{equation}
where $\mathbf r_k$ is the visual residual after eliminating landmark errors, and $\bK_k$ contains the rows of the MSCKF Kalman gain corresponding to the IMU error state $\delta\mathbf x$. The matrix $\mathbf S_k$ is the visual innovation covariance and $\boxplus$ applies the local correction.
The updated IMU state estimate $\hat{\mathbf x}_k^{+}$ and covariance $\bP_k^{+}$ initialize the next propagation step.

\section{Experiments}
\label{sec:experiments}

We evaluate the proposed method on public visual-inertial benchmarks.
Section~\ref{subsec:setup} describes the experimental setup.
Section~\ref{subsec:benchmark} presents visual-inertial estimation results on EuRoC and TUM-VI under nominal conditions.
Section~\ref{subsec:blackout} evaluates visual outage on EuRoC and provides an additional three-dimensional trajectory comparison on TUM-VI.
Section~\ref{subsec:distortion} evaluates visual distortion on EuRoC, and Section~\ref{subsec:noise_ablation} reports ablation studies of the learned bias and uncertainty models.

\subsection{Experimental Setup}
\label{subsec:setup}

\subsubsection{Datasets}
Experiments are conducted on EuRoC MAV and TUM-VI, whose sensor characteristics and data splits are summarized in Table~\ref{tab:datasets}.
EuRoC MAV dataset~\cite{burri2016euroc} comprises 11 indoor sequences acquired in a machine hall and a Vicon room.
The sensor platform records synchronized stereo $752\times480$ global shutter images at 20\,Hz and inertial measurements at 200\,Hz.
\begin{table}[ht]
\centering
\caption{Datasets and train--test splits}
\label{tab:datasets}
\footnotesize
\begin{tabular}{@{}lcccc@{}}
\toprule
Dataset & IMU & Rate & Env. & Train / Test \\ \midrule
EuRoC MAV & ADIS16448 & 200\,Hz & Indoor & 6 / 5 \\
TUM-VI    & BMI160    & 200\,Hz & In/Outdoor & 3 / 3 \\ \bottomrule
\end{tabular}
\end{table}
We use six sequences for training and five for testing.
TUM-VI dataset~\cite{schubert2018tumvi} provides indoor and outdoor sequences recorded with a BMI160 IMU at 200\,Hz.
We use the six indoor room sequences with full ground truth coverage, split into three training and three test sequences.

\subsubsection{Implementation Details}
Each bias dynamics model, $\mathbf f_g$ or $\mathbf f_a$, comprises two residual blocks with Tanh activations and a maximum hidden layer width of 512.
The bias states are integrated at the IMU rate by first order explicit Euler integration with $\Delta t_k=5$\,ms.
Training follows the three stage curriculum described in Sec.~\ref{subsec:training}.
Stages~1 and~2 use $N=16$ sample integration intervals and a batch size of 1000, with each stage trained for 1600 epochs.
\begin{table}[ht]
\centering
\caption{IMU noise density parameters for EuRoC and TUM-VI}
\label{tab:imu_noise}
\footnotesize
\setlength{\tabcolsep}{4.8pt}
\begin{tabular}{@{}lcccc@{}}
\toprule
Parameter & Symbol & EuRoC & TUM-VI & Unit \\ \midrule
Gyro. white noise    & $\sigma_g$       & $1.70\!\times\!10^{-4}$ & $1.60\!\times\!10^{-4}$ & rad/s/$\sqrt{\text{Hz}}$ \\
Accel. white noise   & $\sigma_a$       & $2.00\!\times\!10^{-3}$ & $2.80\!\times\!10^{-3}$ & m/s$^2$/$\sqrt{\text{Hz}}$ \\
Gyro. bias diffusion & $\sigma_{bg}$    & $1.94\!\times\!10^{-5}$ & $2.20\!\times\!10^{-5}$ & rad/s$^2$/$\sqrt{\text{Hz}}$ \\
Accel. bias diffusion & $\sigma_{ba}$   & $3.00\!\times\!10^{-3}$ & $8.60\!\times\!10^{-4}$ & m/s$^3$/$\sqrt{\text{Hz}}$ \\ \bottomrule
\end{tabular}
\end{table}

\begin{table*}[tp]
\caption{VIO accuracy on the EuRoC and TUM-VI test sequences under nominal visual conditions}
\label{tab:vio}
\centering
\footnotesize
\setlength{\tabcolsep}{10pt}
\begin{tabular}{@{}ll cc cc cc cc cc@{}}
\toprule
 & & \multicolumn{2}{c}{S-MSCKF~\cite{sun2018robust}} & \multicolumn{2}{c}{MSCEqF~\cite{fornasier2023msceqf}} & \multicolumn{2}{c}{VIO-IPNet~\cite{yi2026plug}} & \multicolumn{2}{c}{Brossard~\cite{brossard2020denoising}} & \multicolumn{2}{c}{LBDU-VIO} \\ 
\cmidrule(lr){3-4}\cmidrule(lr){5-6}\cmidrule(lr){7-8}\cmidrule(lr){9-10}\cmidrule(lr){11-12}
Dataset & Sequence & APE & AOE & APE & AOE & APE & AOE & APE & AOE & APE & AOE \\
\midrule
\multirow{6}{*}{EuRoC}
 & MH\_02 & \textbf{0.141} & 1.860 & 0.639 & 3.117 & 0.166 & \textbf{1.721} & 0.196 & 2.334 & 0.232 & 2.221 \\
 & MH\_04 & 0.355 & 2.243 & 0.499 & 1.818 & 0.325 & 1.790 & 0.300 & 1.701 & \textbf{0.215} & \textbf{1.194} \\
 & V1\_01 & 0.090 & 1.276 & 0.132 & 6.185 & 0.086 & 0.938 & 0.098 & 1.151 & \textbf{0.055} & \textbf{0.470} \\
 & V1\_03 & 0.247 & 6.911 & 0.167 & \textbf{2.106} & 0.095 & 2.514 & 0.093 & 2.299 & \textbf{0.089} & 3.155 \\
 & V2\_02 & 0.163 & 3.097 & 0.208 & 3.149 & 0.098 & 1.839 & 0.129 & 1.955 & \textbf{0.076} & \textbf{1.412} \\
\cmidrule(lr){2-12}
 & Mean   & 0.199 & 3.077 & 0.329 & 3.275 & 0.154 & 1.760 & 0.163 & 1.888 & \textbf{0.133} & \textbf{1.690} \\
\midrule
\multirow{4}{*}{TUM-VI}
 & room2  & 0.144 & 6.296 & 0.224 & 2.879 & 0.094 & 2.259 & 0.094 & 2.407 & \textbf{0.088} & \textbf{2.187} \\
 & room4  & 0.096 & 3.165 & 0.270 & 6.080 & 0.041 & 1.969 & 0.522 & 4.730 & \textbf{0.028} & \textbf{1.116} \\
 & room6  & 0.140 & 2.078 & 0.211 & 3.435 & 0.101 & 2.222 & 0.072 & 2.551 & \textbf{0.064} & \textbf{1.651} \\
\cmidrule(lr){2-12}
 & Mean   & 0.127 & 3.846 & 0.235 & 4.131 & 0.079 & 2.150 & 0.229 & 3.229 & \textbf{0.060} & \textbf{1.651} \\
\bottomrule
\end{tabular}
\end{table*}
Stage~1 learns $\mathbf f_g$ from attitude supervision.
Stage~2 freezes $\mathbf f_g$ and learns $\mathbf f_a$ from velocity and position supervision.
The curvature regularization weights in \eqref{eq:stage_loss_rotation} and
\eqref{eq:stage_loss_velocity_position} are set to $\lambda_g=10^{-10}$ and $\lambda_a=10^{-12}$, respectively.
Both stages use Adam with a learning rate of $5\!\times\!10^{-3}$ on one NVIDIA RTX~4060 Laptop GPU.
In Stage~3, we train the uncertainty model using 40 sample windows and a batch size of 64 with Adam. It is initialized at a learning rate of $10^{-3}$ and trained for at most 500 epochs.
We next describe the estimator configuration.
The estimator maintains up to $W=10$ camera poses and performs visual updates at 20\,Hz on both datasets.
Table~\ref{tab:imu_noise} lists the manufacturer specified continuous time noise densities.
Without the uncertainty model, the measurement noise covariances are fixed as $\bQ_{ng}=\sigma_g^2\I_3$ and $\bQ_{na}=\sigma_a^2\I_3$.
The uncertainty model replaces these matrices with $\bQ_{ng,k}$ and $\bQ_{na,k}$, while the bias process covariances $\bQ_{bg}=\sigma_{bg}^2\I_3$ and $\bQ_{ba}=\sigma_{ba}^2\I_3$ remain fixed at their datasheet values.

\subsubsection{Evaluation Metrics}
We assess global trajectory accuracy and local drift using absolute and relative errors, respectively.
For global accuracy, the estimated trajectory is first aligned to the ground truth in $\mathrm{SE}(3)$~\cite{sola2018lie}.
We then report the root mean square angular and position errors as the absolute orientation error (AOE) and absolute position error (APE)
\begin{equation}
\begin{aligned}
\mathrm{AOE}&=\sqrt{\frac{1}{N}\sum_{k=1}^{N}\lVert\Log(\hat{\R}_k^T\R_k^{\mathrm{gt}})\rVert_2^2}\\
\mathrm{APE}&=\sqrt{\frac{1}{N}\sum_{k=1}^{N}\lVert\hat{\p}_k-\p_k^{\mathrm{gt}}\rVert_2^2}
\end{aligned}
\label{eq:abs}
\end{equation}
Here, $N$ is the number of associated poses. AOE and ROE are reported in degrees.
We assess local trajectory drift over fixed length path segments using the relative orientation error (ROE) and relative position error (RPE), defined as
\begin{equation}
\mathrm{ROE}=\frac{1}{D}\sum_{s=1}^{D}\left\|\Log\!\left(\R_{e,s}\right)\right\|_2, \,
\mathrm{RPE}=\frac{1}{D}\sum_{s=1}^{D}\left\|\p_{e,s}\right\|_2
\label{eq:rel}
\end{equation}
Here, $D$ is the number of evaluated ground truth segments, each with a fixed path length $d$.
For segment $s$, $\R_{e,s}$ and $\p_{e,s}$ are the rotation and translation components of the relative pose error $\mathbf T_{e,s}=\Delta\mathbf T_s^{-1}\Delta\hat{\mathbf T}_s$.
The increments $\Delta\mathbf T_s$ and $\Delta\hat{\mathbf T}_s$ represent the ground truth and estimated relative poses in $\SE(3)$ over the same segment, respectively.
APE and RPE are reported in meters.

\subsubsection{Comparison Methods}

We evaluate the proposed integrated VIO system against four representative baselines.
Among the filter-based baselines, S-MSCKF~\cite{sun2018robust} uses synchronized stereo feature tracks in MSCKF updates to improve robustness while retaining computational efficiency.
MSCEqF~\cite{fornasier2023msceqf} is a monocular equivariant filter that embeds navigation and camera calibration states in a unified symmetry group to improve consistency under large estimation errors.
Among the learning-based baselines, VIO-IPNet~\cite{yi2026plug} restructures the IMU bias prediction paradigm and proposes an end-to-end average bias regression network, whereas Brossard~\emph{et al.}~\cite{brossard2020denoising} denoise the gyroscope stream using a dilated convolutional network.
Both learned baselines are retrained on the corresponding EuRoC and TUM-VI training splits and coupled with unmodified OpenVINS framework.
MSCEqF and S-MSCKF are evaluated using their released implementations, with S-MSCKF operating in stereo.
The implementation of our proposed system extends the error state Kalman filter with learned bias dynamics, an uncertainty model, and the standard MSCKF visual updates.
The common platform comparisons control differences in visual processing.

\subsection{VIO Evaluation Under Nominal Conditions}
\label{subsec:benchmark}

This experiment evaluates the proposed method accuracy under continuous $20$\,Hz visual updates, with the learned bias dynamics and inertial measurement uncertainty modules.
Table~\ref{tab:vio} reports APE and AOE on EuRoC and TUM-VI, where lower values indicate better accuracy.
All systems use monocular input except the stereo S-MSCKF.
On EuRoC, the complete proposed system ranks first in both overall metrics, with $0.133$\,m mean APE and $1.690^{\circ}$ mean AOE.
Relative to VIO-IPNet, the strongest baseline for both metrics, it reduces mean APE by $13.6\%$ and mean AOE by $4.0\%$.
This advantage is also reflected at the sequence level, with the lowest APE on four of five sequences and the lowest AOE on three.
MH\_04 sequence provides a representative joint gain, as APE and AOE decrease by $28.3\%$ and $29.8\%$, respectively, against the best baseline for each metric.
However, the sequence level gains do not hold uniformly across both metrics.
On MH\_02, S-MSCKF retains the lowest APE, whereas VIO-IPNet retains the lowest AOE.
On V1\_03, the proposed system leads APE, but its AOE remains $49.8\%$ above the MSCEqF minimum.
Table~\ref{tab:vio} therefore supports an overall system level advantage rather than superiority across all sequences.
\begin{figure}[!t]
\centering
\includegraphics[width=0.92\columnwidth]{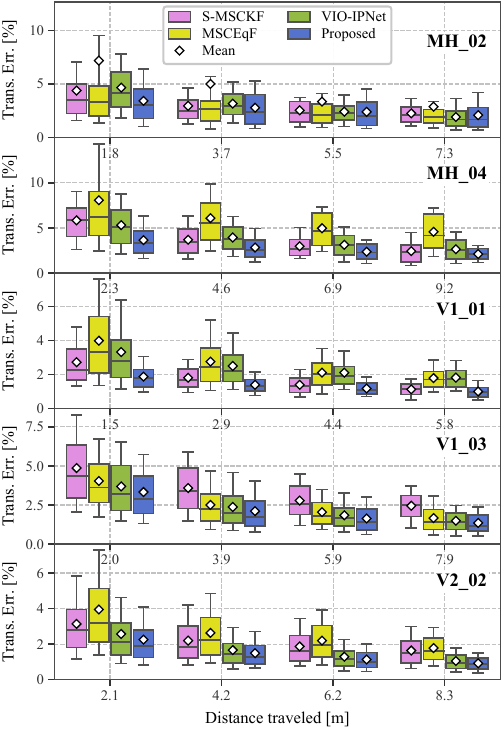}
\caption{Relative translation drift on the five EuRoC test sequences.}
\label{fig:rpe}
\end{figure}
Local translational drift is evaluated under the same nominal visual conditions to complement the absolute errors, as shown in Fig.~\ref{fig:rpe}.
The relative translational drift percentage plotted on the vertical axis is $e_{t,s}(d)=\frac{\|\p_{e,s}\|_2}{d}\times100\%$.
We compute $e_{t,s}(d)$ for all sub trajectories of length $d$ and the boxplot shows the statistical distribution of these errors.
For each sequence, the target sub-trajectory length $d$ is set to $2.5\%$, $5\%$, $7.5\%$, and $10\%$ of the total ground truth path length, with the corresponding distances shown on the horizontal axis.
Each box spans the 25th to 75th percentiles, with whiskers extending to the 10th and 90th percentiles.

Across four more complexed path lengths, the proposed system achieves the lowest mean RPE among all methods at each evaluated path length.
MH\_02 forms the exception, where the proposed system leads at the two shorter lengths and VIO-IPNet leads at the two longer lengths.
The relatively mild motion in this sequence may limit the benefit of nonlinear bias learning.
The local drift advantage therefore covers most sequences but weakens as path length increases on MH\_02.
On TUM-VI, the proposed monocular system ranks first in both metrics on room2, room4, and room6.
Relative to VIO-IPNet, the closest monocular baseline with complete results, it reduces mean APE by $24.1\%$ and mean AOE by $23.2\%$.
The proposed system lowers mean APE by $74.5\%$ and mean AOE by $60.0\%$ compared with MSCEqF.
Its advantage over Brossard is similarly clear, with mean APE and AOE reduced by $73.8\%$ and $48.9\%$, respectively, while room4 APE falls from $0.522$\,m to $0.028$\,m.
These results demonstrate the effectiveness of jointly integrating the learned bias and uncertainty modules within the proposed method.
Despite using monocular input, the proposed system reduces mean APE and AOE by $52.8\%$ and $57.1\%$ relative to the stereo S-MSCKF.

\subsection{Robustness Under Visual Outage}
\label{subsec:blackout}

\subsubsection{Effect of Outage Duration on EuRoC}
This experiment examines whether the proposed method limits drift during a temporary loss of vision and remains accurate after visual measurements return.
We insert one outage into each of the five EuRoC test sequences.
The outage starts $50$\,s after the beginning of the sequence and lasts $1$, $2$, $3$, $5$, or $10$\,s.
Every method is evaluated with the same outage start and end times.
During the outage, all camera messages used by that method are removed, while the IMU stream is left unchanged.
When the outage ends, visual measurements resume and the estimator continues to the end of the sequence.
This protocol tests both IMU-only propagation during the outage and the system behavior after vision returns.
\begin{table}[!t]
\centering
\caption{Mean RPE and ROE across the five EuRoC test sequences at each visual-outage duration. The best baseline is selected separately for each metric and duration. Best values are shown in bold.}
\label{tab:outage_duration_summary}
\footnotesize
\setlength{\tabcolsep}{4pt}
\begin{tabular}{@{}l ccc ccc@{}}
\toprule
\multirow{2}{*}{Outage}
  & \multicolumn{3}{c}{RPE (m)}
  & \multicolumn{3}{c}{ROE ($^\circ$)} \\
\cmidrule(lr){2-4}\cmidrule(lr){5-7}
  & \shortstack{Best\\Baseline} & Proposed & Reduction
  & \shortstack{Best\\Baseline} & Proposed & Reduction \\
\midrule
1\,s  & 0.052 & \textbf{0.044} & 15.4\% & 0.304 & \textbf{0.275} & 9.5\% \\
2\,s  & 0.053 & \textbf{0.046} & 13.2\% & 0.308 & \textbf{0.273} & 11.4\% \\
3\,s  & 0.060 & \textbf{0.044} & 26.7\% & 0.308 & \textbf{0.254} & 17.5\% \\
5\,s  & 0.079 & \textbf{0.058} & 26.6\% & 0.306 & \textbf{0.259} & 15.4\% \\
10\,s & 0.187 & \textbf{0.140} & 25.1\% & 0.320 & \textbf{0.306} & 4.4\% \\
\bottomrule
\end{tabular}
\end{table}

\begin{figure}[!t]
\centering
\includegraphics[width=\columnwidth]{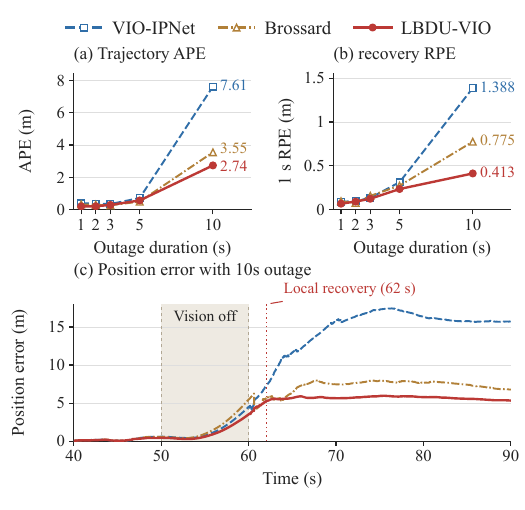}
\caption{Visual outage and recovery on EuRoC MH\_04\_difficult sequence.
(a) Trajectory APE after global $\SE(3)$ alignment.
(b) Recovery RPE denotes $1$\,s relative translation RMSE over the first $10$\,s after images resume.
(c) Position error with pre outage alignment.
Shading marks the $50$-$60$\,s outage and red dotted lines indicate local recovery near $62$\,s.}
\label{fig:outage}
\end{figure}

\begin{figure}[!t]
\centering
\includegraphics[width=\columnwidth]{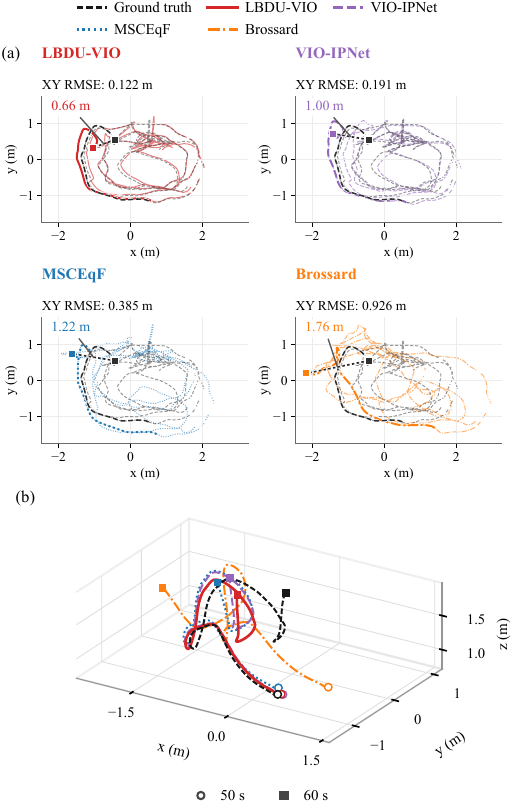}
\caption{Trajectories on TUM-VI room4 with a $10$\,s visual outage. (a) XY projections over $20$-$110$\,s, with planar RMSE reported above each plot. Thick segments show the outage, and connected squares mark XY errors at $60$\,s. (b) Three dimensional trajectories during the outage. Circles and squares mark $50$ and $60$\,s. Each method shows the run with median APE among three repeats.}
\label{fig:room4_3d_outage}
\end{figure}

Table~\ref{tab:outage_duration_summary} reports the mean RPE and ROE across the five sequences at each outage duration.
Both metrics are computed over fixed $1$\,m path increments on the complete output trajectory.
They therefore reflect the combined effects of drift during the outage and estimation after visual updates resume.
The best baseline is selected separately for each duration and metric as the baseline with the lowest across sequence mean.
For RPE, this baseline is VIO-IPNet at $1$-$3$\,s and S-MSCKF at $5$-$10$\,s. For ROE, it is MSCEqF at every duration.
The summary shows a consistent overall advantage for the proposed system.
Proposed achieves the lowest across-sequence mean RPE and ROE at every tested outage duration.
Relative to the best baseline in each condition, the RPE reduction ranges from $13.2\%$ to $26.7\%$, and the ROE reduction ranges from $4.4\%$ to $17.5\%$.
Because the reference baseline is chosen independently in each column, these reductions are measured against the strongest baseline result for that duration and metric.
The size of the gain changes with outage duration.
At $1$-$2$\,s, Proposed reduces mean RPE by $15.4\%$ and $13.2\%$, and mean ROE by $9.5\%$ and $11.4\%$, respectively.
The joint translation and rotation advantage is largest at $3$-$5$\,s. The RPE reduction is $26.7\%$ and $26.6\%$, while the ROE reduction is $17.5\%$ and $15.4\%$.
At $10$\,s, mean RPE increases to $0.140$\,m, but remains $25.1\%$ below the best baseline value of $0.187$\,m.
In contrast, the ROE margin narrows to $4.4\%$, with $0.306^{\circ}$ for Proposed and $0.320^{\circ}$ for the best baseline.
The translational advantage therefore remains clear at the longest outage, whereas the smaller orientation margin shows that long IMU-only propagation remains challenging.

Fig.~\ref{fig:outage} compares LBDU-VIO, VIO-IPNet, and Brossard across five outage durations on the EuRoC MH\_04 sequence.
In Fig.~\ref{fig:outage}(a), LBDU-VIO achieves the lowest trajectory APE at $1$, $2$, $5$, and $10$\,s, while Brossard has lower APE at $3$\,s.
For the $10$\,s outage, LBDU-VIO achieves an APE of $2.74$\,m, compared with $3.55$\,m for Brossard and $7.61$\,m for VIO-IPNet.
This corresponds to a $22.9\%$ reduction relative to the stronger baseline.
Fig.~\ref{fig:outage}(b) complements trajectory APE by measuring relative translation accuracy during the first $10$\,s after images resume.
LBDU-VIO achieves the lowest recovery RPE at four of the five outage durations, with Brossard leading at $2$\,s.
For the $10$\,s outage, its $1$\,s recovery RPE is $0.413$\,m, compared with $0.775$\,m for Brossard and $1.388$\,m for VIO-IPNet.
The $46.7\%$ reduction relative to Brossard shows improved relative motion accuracy alongside the lower trajectory APE.
Fig.~\ref{fig:outage}(c) shows the position error for the $50$--$60$\,s outage.
All three methods exhibit small position errors before the outage, followed by increasing errors during IMU only propagation.
By the end of the outage, LBDU-VIO has the lowest position error among the three methods.
After images resume, its position error remains approximately $5$--$6$\,m from $63$ to $90$\,s.
At $90$\,s, LBDU-VIO has a position error of $5.33$\,m, compared with $6.78$\,m for Brossard and $15.77$\,m for VIO-IPNet.
These results show lower accumulated position error at the end of the outage and limited further error growth after visual updates resume.

\subsubsection{Trajectory Comparison on TUM-VI}
We compare LBDU-VIO, VIO-IPNet, MSCEqF, and Brossard on TUM-VI room4 using monocular images, with three runs per method.
We remove camera measurements during $50$-$60$\,s and retain all IMU samples.
We compute translational APE after one global $\SE(3)$ alignment per run, using ground truth timestamps over $20$-$110$\,s.
Fig.~\ref{fig:room4_3d_outage} shows XY projections and three dimensional outage trajectories for the run with median APE from each method.
In the XY view, LBDU-VIO follows the reference path more closely than the baselines.
Just before visual updates resume, its XY position error is $0.66$\,m, compared with $1.00$\,m for VIO-IPNet.
The corresponding errors for MSCEqF and Brossard are $1.22$\,m and $1.76$\,m, respectively.
LBDU-VIO achieves the lowest mean APE of $0.1349$\,m, compared with $0.1920$\,m for VIO-IPNet, $0.4316$\,m for MSCEqF, and $1.2348$\,m for Brossard.
This is a $29.8\%$ reduction relative to VIO-IPNet, the strongest baseline in this experiment.
LBDU-VIO also achieves lower APE than all three baselines in every run.

\subsection{Robustness Under Visual Distortion}
\label{subsec:distortion}

This experiment compares estimation accuracy under repeated visual distortion.
We use the EuRoC {V2\_02\_medium} sequence and apply a horizontal pixel shift ramp to the monocular {cam0} stream.
The distortion starts at $15$\,s and lasts for $5$\,s every $15$\,s, leaving a $10$\,s normal image interval between consecutive windows.
Within each window, the shift increases linearly from zero to $10$\,px.
For a horizontal focal length of $458.654$\,px, the maximum shift corresponds to an angular displacement of approximately $1.25^\circ$.
We compare LBDU-VIO with MSCEqF, Brossard, and VIO-IPNet using the same distorted monocular images and IMU measurements for all methods.
We run each method three times under visual distortion.
All trajectories are restricted to their common time interval, from $4$ to $116$\,s.
Each trajectory is then independently aligned to the ground truth with one $\SE(3)$ transform estimated over this interval.

\begin{table}[!t]
\centering
\caption{Estimation accuracy under intermittent visual distortion on EuRoC {V2\_02\_medium}. Values are median RMSEs and bold indicates the lowest value in each column.}
\label{tab:distortion_sensitivity}
\setlength{\tabcolsep}{4pt}
\renewcommand{\arraystretch}{1.08}
\footnotesize
\begin{tabular}{lcc}
\toprule
Method & AOE RMSE ($^\circ$) & APE RMSE (m) \\
\midrule
MSCEqF   & 2.619 & 0.175 \\
Brossard & 1.779 & 0.194 \\
VIO-IPNet & 2.181 & 0.152 \\
\textbf{LBDU-VIO} & \textbf{1.378} & \textbf{0.101} \\
\bottomrule
\end{tabular}%
\end{table}

\begin{figure}[!t]
\centering
\includegraphics[width=\columnwidth]{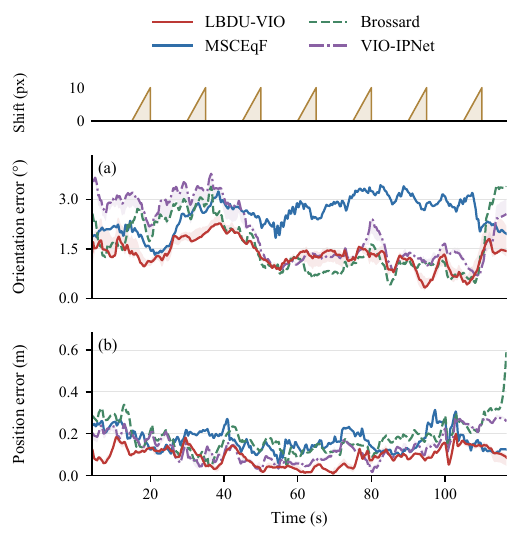}
\caption{Orientation (a) and position (b) errors under intermittent visual distortion on EuRoC {V2\_02\_medium}. The upper track shows the imposed horizontal image shift. Lines show the median error at each timestamp, and shaded bands span the minimum and maximum errors across three runs.}
\label{fig:distortion}
\end{figure}
\begin{figure*}[!t]
\centering
\includegraphics[width=\textwidth]{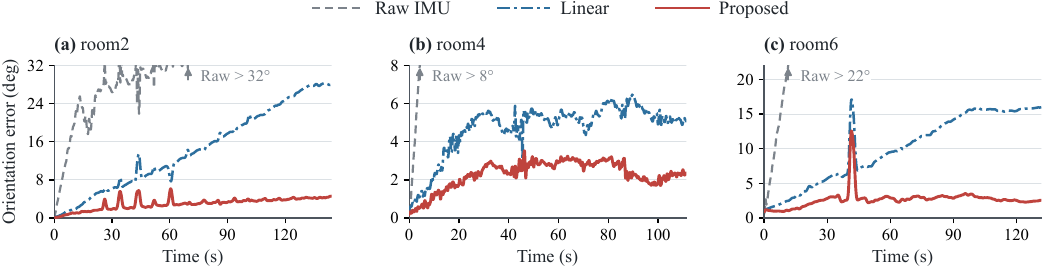}
\caption{Orientation errors during open loop gyroscope integration on the TUM-VI test sequences: (a) {room2}, (b) {room4}, and (c) {room6}. Larger Raw IMU errors are clipped at the upper boundary, and the upward arrow shows that this trace continues beyond the displayed range.}
\label{fig:tum_pureint}
\end{figure*}
\begin{table*}[!t]
\centering
\caption{Component ablation under a three-second visual outage on the EuRoC test sequences. Values are reported as APE/RPE in meters. Lower values indicate better accuracy, and the best results are shown in bold.}
\label{tab:noise_ablation}
\footnotesize
\begin{tabular}{l ccccc c}
\toprule
\multirow{2}{*}{Method}
  & MH02 & MH04 & V103 & V202 & V101 & Mean \\
  & \multicolumn{6}{c}{\footnotesize APE\,/\,RPE\,(m)} \\
\midrule
Raw IMU + Fixed Noise
  & 0.153\,/\,0.087 & 0.306\,/\,0.104 & 0.085\,/\,0.075 & 0.606\,/\,0.158 & 0.097\,/\,0.093 & 0.249\,/\,0.103 \\
Learned Bias + Fixed Noise
  & 0.145\,/\,0.055 & 0.284\,/\,0.078 & 0.085\,/\,0.059 & 0.150\,/\,0.057 & 0.127\,/\,0.094 & 0.158\,/\,0.069 \\
Raw IMU + uncertainty model
  & 0.160\,/\,0.088 & 0.296\,/\,0.107 & 0.095\,/\,0.074 & 0.557\,/\,0.133 & 0.090\,/\,0.089 & 0.240\,/\,0.098 \\
Learned Bias + uncertainty model
  & \textbf{0.107}\,/\,\textbf{0.050} & \textbf{0.227}\,/\,\textbf{0.060}
  & \textbf{0.077}\,/\,\textbf{0.050} & \textbf{0.088}\,/\,\textbf{0.039}
  & \textbf{0.054}\,/\,\textbf{0.024} & \textbf{0.111}\,/\,\textbf{0.045} \\
\bottomrule
\end{tabular}%
\end{table*}

Table~\ref{tab:distortion_sensitivity} summarizes estimation accuracy under visual distortion.
LBDU-VIO achieves the lowest median AOE and APE RMSEs across three runs.
Under visual distortion, its median AOE RMSE is $1.378^\circ$, compared with $1.779^\circ$ for Brossard. This corresponds to a $22.5\%$ reduction relative to the best orientation baseline.
Its median APE RMSE is $0.101$\,m, compared with $0.152$\,m for VIO-IPNet. This corresponds to a $33.3\%$ reduction relative to the best position baseline.
Fig.~\ref{fig:distortion} shows how orientation and position errors evolve across repeated distortion cycles.
LBDU-VIO maintains lower orientation errors than MSCEqF, although Brossard and VIO-IPNet have lower errors at some timestamps.
Its position error is lower than those of all three baselines for most of the sequence.
Across the three runs, LBDU-VIO achieves a maximum orientation error of $2.524^\circ$, compared with $3.455^\circ$ for MSCEqF, the best baseline on this metric.
Its maximum position error is $0.212$\,m, compared with $0.297$\,m for VIO-IPNet, the best baseline on this metric.

\subsection{Evaluation and Ablation of the Learned Components}
\label{subsec:noise_ablation}

We evaluate the learned components in stages, beginning with the bias model in isolation.
To exclude the effects of visual updates, filter correction, and learned uncertainty, Fig.~\ref{fig:tum_pureint} integrates each gyroscope stream in open loop from the ground truth initial attitude.
The comparison uses Raw IMU measurements and streams corrected by the Linear baseline and Proposed bias model.
The Linear baseline computes the calibrated angular velocity as $\bar{\boldsymbol\omega}_k=\mathbf A_g\tilde{\boldsymbol\omega}_k+\mathbf c_g$ at each IMU sample. 
We learn the calibration matrix $\mathbf A_g\in\mathbb R^{3\times3}$ and offset $\mathbf c_g\in\mathbb R^3$ on the training sequences by minimizing orientation error after integration.
Here, Proposed denotes the learned bias model alone rather than the proposed method.
To keep the Linear and Proposed traces fully visible, the vertical axis spans $0$-$8^{\circ}$.
On {room4} sequence, Proposed maintains the orientation error near $2^\circ$, with a mean of $2.3^\circ$, substantially lower than that of the Raw IMU. In comparison, the Linear model remains near $5^\circ$. Moreover, Proposed consistently outperforms Linear throughout the sequence, indicating that the improvement is sustained rather than limited to a short interval.
The same ordering holds on {room2} and {room6}. It indicates that the learned bias model consistently suppresses orientation drift during open loop integration across all three test sequences.
Having examined the learned bias model in isolation, we next assess its interaction with the uncertainty model under visual outage. Table~\ref{tab:noise_ablation} summarizes the separate and joint evaluations of the learned bias model and uncertainty model under a 3\,s visual outage on the five EuRoC test sequences. Each cell reports APE followed by RPE, both in meters. RPE is computed over fixed 1\,m path segments, and the Mean column averages the five displayed sequences. The learned bias model propagates the filter's bias state, which is used to correct IMU measurements for navigation propagation. The uncertainty model supplies measurement noise covariances for covariance propagation. The final row enables both modules and represents the complete proposed system.

To isolate the contribution of learned bias dynamics under fixed measurement noise, we compare \emph{Raw IMU + Fixed Noise} with \emph{Learned Bias + Fixed Noise} in Table~\ref{tab:noise_ablation}. Adding learned bias reduces the mean APE from $0.249$ to $0.158$\,m. This corresponds to a reduction of $36.5\%$. The mean RPE decreases from $0.103$ to $0.069$\,m. This corresponds to a reduction of $33.0\%$. Across the five sequences, APE decreases in three cases. It remains unchanged in one case. RPE decreases in four cases. The largest reductions are observed on V2\_02. Its APE decreases from $0.606$ to $0.150$\,m, a reduction of $75.2\%$. Its RPE decreases from $0.158$ to $0.057$\,m, a reduction of $63.9\%$. 
To isolate the contribution of the uncertainty model, we hold the learned bias correction fixed and compare \emph{Learned Bias + Fixed Noise} with \emph{Learned Bias + uncertainty model}. Adding the uncertainty model reduces mean APE from $0.158$ to $0.111$\,m and mean RPE from $0.069$ to $0.045$\,m, corresponding to reductions of $29.7\%$ and $34.8\%$, respectively. Both metrics decrease on all five sequences, and the complete configuration achieves the lowest APE and RPE in every reported column.
By contrast, adding the uncertainty model to the raw IMU changes mean APE only from $0.249$ to $0.240$\,m and mean RPE from $0.103$ to $0.098$\,m. This interaction is consistent with AirIMU~\cite{qiu2024airimu}, where learned covariance is not uniformly beneficial with raw IMU inputs but performs best when combined with IMU correction. Together, these results indicate that mean correction and covariance modeling provide complementary benefits.


\section{Conclusion}
\label{sec:conclusion}

In this paper, we present LBDU-VIO, a tightly coupled VIO framework that augments MSCKF with learned continuous time IMU bias dynamics and adaptive IMU uncertainty model.
The bias model corrects IMU rate state propagation, while the uncertainty model supplies measurement noise covariances at every IMU sample between visual updates.
The two modules are trained hierarchically without direct ground truth labels.
Experiments on the real world EuRoC and TUM-VI benchmarks show consistent accuracy gains over representative baselines, especially during visual outages when the estimator relies mainly on IMU measurements.
These results suggest that learning to refine inertial propagation offers a practical path toward robust VIO. 
Future work focuses on generalization across longer outages and less structured visual degradation.

\begingroup
\footnotesize
\bibliographystyle{IEEEtranN}
\bibliography{customabrv,IEEEabrv,paper}
\endgroup

\vspace{12pt}

\end{document}